\documentclass[11pt,a4paper]{article}
\pdfoutput=1
\usepackage[hyperref]{emnlp2020}
\usepackage{times}
\usepackage{latexsym}

\usepackage{microtype}

\aclfinalcopy % Uncomment this line for the final submission
\def\aclpaperid{6} %  Enter the acl Paper ID here

\usepackage{hhline}
\usepackage{multirow}
\usepackage[frozencache,cachedir=.]{minted}

\title{iobes: A Library for Span-Level Processing}

\author{
Brian Lester \\
Independent \\
blester125@gmail.com
}

\date{}

\begin{document}
\maketitle
\begin{abstract}
Many tasks in natural language processing, such as named entity recognition and slot-filling, involve identifying and labeling specific spans of text. In order to leverage common models, these tasks are often recast as sequence labeling tasks. Each token is given a label and these labels are prefixed with special tokens such as \texttt{B-} or \texttt{I-}. After a model assigns labels to each token, these prefixes are used to group the tokens into spans.

Properly parsing these annotations is critical for producing fair and comparable metrics; however, despite its importance, there is not an easy-to-use, standardized, programmatically integratable library to help work with span labeling. To remedy this, we introduce our open-source library, \textit{iobes}. \textit{iobes} is used for parsing, converting, and processing spans represented as token level decisions. 
\end{abstract}

\section{Introduction}

Tasks like named entity recognition, finding mentions for real world things in text, and slot-filling, finding mentions of relevant objects, often in a dialogue, require identifying contiguous sections of the input text and classifying them into one of several pre-defined classes. While some work solves span labeling by scoring all possible spans, followed by filtering with a threshold \cite{lee-etal-2017-end}, most work recasts span identification as a token labeling task \cite{tjong-kim-sang-de-meulder-2003-introduction,bender-etal-2003-maximum,JMLR:v12:collobert11a,ma-hovy-2016-end,lample-etal-2016-neural}. Special prefixes like \texttt{B-} are used in conjunction with token level type labels, like ``PER'' for person or ``ORG'' for organization, to signal where different spans begin and end. Once a standard sequence labeling model is used to produce tags for each token, we use these prefixes to convert the token level annotations into spans. For example, the token \texttt{B-LOC} means that we need to start a location span here. The tokens labeled with \texttt{I-LOC} are continuations of this span. Finally, a token of a different type, or the special \texttt{O} label, representing a token that is ``outside'' of a span, will signal the end of our span. Once we have decoded our spans, we often use a metric like exact match F1, where both the span type and the span boundaries have to match, to compare our predicted spans to the reference spans.

Unfortunately, there are many subtle places where implementations of this approach to span labeling can diverge. The semantics of these special prefixes can change. There are multiple encoding schemes that are all equally expressive but have differences in how easy it is for a model to learn them \cite{ratinov-roth-2009-design}. Many common datasets are old and therefore distributed in older formats. Researchers often convert these datasets to newer formats, but differences or bugs in this process can introduce discrepancies in the data used to train models. Policy decisions on handling token level annotations that do not conform to the rules of the encoding scheme also introduce differences that render different models incomparable.

The problems mentioned above all stem from the lack of a community wide standardization on how to handle these edge cases. The lack of common tooling for working with these spans, represented as token level annotations, has led to many rolling their own, slightly different implementations. We introduce the \textit{iobes} library. A Python \cite{10.5555/1593511} library that encapsulates all the rough edges of processing and evaluating spans. We aim to provide the community with a single, easy-to-use toolkit whose adoption will ensure the comparability of span labeling metrics reported by different researchers.

\section{Formats}

There are several forms that span encoding via token labeling can take. Later we will see how these multiple formats---and the conversion between them---are the cause of many problems, but for now we will summarize them here.

\begin{itemize}
    \item \textbf{IOB}: The original span labeling format introduced by \citet{ramshaw-marcus-1995-text}. In this format, tokens that are not part of a span are labeled with \texttt{O}. Tokens that are part of a span are tagged with the span type, prefixed with a \texttt{I-}; for example, \texttt{I-PER} is part of a span representing a person. The special \texttt{B-} prefix is used to demarcate two spans of the same type that touch.
    \item \textbf{BIO}: A simple extension of the IOB format where all entities, regardless of what entities they touch, begin with a \texttt{B-}. An advantage of this format is that it is no longer contextual. The span ``Real Madrid'' will always have the tags \texttt{B-ORG I-ORG}, regardless of what the previous span is. In the IOB scheme, this would be \texttt{I-ORG I-ORG} by default and would only use the \texttt{B-} tag when preceded by another span of type ``ORG''.
    \item \textbf{IOBES}: A further extension to the BIO labeling scheme. It adds two new tags types. \texttt{E-} is used to label the token that is the last item of a span. The new \texttt{S-} prefix is used for span that only include a single token. In our example the span ``Real Madrid'' would be labeled as \texttt{B-ORG E-ORG}. This span encoding format has several names. BILOU is the same scheme, but uses a \texttt{L-} instead of \texttt{E-} for span ending tokens and uses \texttt{U-} rather than \texttt{S-} for single token spans. There is also the BMEWO format. This was the original name for the format introduced in \citet{10.5555/930095}. It uses \texttt{W-} in place of \texttt{S-} and it actually replaces the tokens inside of the spans, using \texttt{M-} meaning middle over \texttt{I-} for inside. This format has be demonstrated to yield better performing models \cite{ratinov-roth-2009-design}.
\end{itemize}

\section{Discrepancies}

Differences in how processing of these spans is done can cause discrepancies in both the datasets and the evaluation metrics used by researchers. While this section draws examples of errors from specific pieces of work, we would like to emphasize these are not failures on the part of the authors, but rather a failing of the community for not providing tested, reusable tooling.

\subsection{Conversion}

Many older datasets like CONLL 2003 \cite{tjong-kim-sang-de-meulder-2003-introduction} are distributed in older formats such as IOB. Researchers then convert them to newer formats like IOBES. This conversion can go wrong. For example, the data used in \citet{yang-etal-2018-design} contained such errors. A Github issue\footnote{\href{https://github.com/jiesutd/NCRFpp/issues/36}{https://github.com/jiesutd/NCRFpp/issues/36}} points out that two entities, one of length one followed by one of length two, in the original IOB1 format (represented by the tag sequence \texttt{I-MISC B-MISC I-MISC}) were incorrectly converted into three separate, length-one entities. The author states that this transformed data was received from a friend, meaning that there is probably another paper---that likely did not open source their data and code---that has this same error. The authors analysis claims that the number of these mistakes are too small to affect the results when the metric is aggregated over the whole test sets. Regardless of whether this bug affected these particular results or not, it is worrying that different researchers are using different datasets. After all, the point of shared datasets is to hold the input data constant, allowing one to demonstrate the improvements are truly from their new modeling approach.

\subsection{Formatting}

A second place where errors can creep in is the formatting of the token level annotations. Different encoding schemes have different rules, for example, in the BIO tagging scheme each entitiy needs to start with a \texttt{B-} tag. This means that \texttt{I-} tokens can only follow \texttt{B-} tokens of the same type. In the data set for the WNUT 2017 shared task \cite{derczynski-etal-2017-results}, there was an entity of type ``creative-work'' that incorrectly started with an \texttt{I-} token. While this has since been fixed\footnote{\href{https://github.com/leondz/emerging\_entities\_17/pull/4}{https://github.com/leondz/emerging\_entities\_17/pull/4}}, gold data that does not strictly follow the rules of the encoding scheme puts toolkits that provide tagger output modules that enforce constraints based on the tagging scheme, such as AllenNLP \cite{Gardner2017AllenNLP} and Mead-Baseline \cite{W18-2506}, at a disadvantage. If you constrain your output to follow the encoding scheme, but the answer does not follow it, your model literally cannot get this example correct. Errors such as these can also cause problems for encoding scheme conversion code, which often assumes that the input is well-formed. Different error handling policies will create different gold data.

\begin{table*}[th!]
    \centering
    \begin{tabular}{c|lll|rrr}
         Index & Surface  & Gold Tag       & Predicted Tag   & Gold      & NCRF++      & \texttt{conlleval.pl} \\
         \hhline{=|===|===}
         0     & to       & \texttt{O}     & \texttt{O}      &           &             &       \\
         1     & First    & \texttt{B-ORG} & \texttt{B-ORG}  &           &             & ORG @ 1 \\ 
         2     & National & \texttt{I-ORG} & \texttt{I-MISC} & ORG @ 1-3 & ORG @ 1-3   & MISC @ 2 \\
         3     & Bank     & \texttt{E-ORG} & \texttt{E-ORG}  &           &             & ORG @ 3 \\
    \end{tabular}
    \caption{
        Differences in the handling of the malformed tags can yield different entities. The gold annotation is a span of type organization starting at index 1 and continuing until index 3 (inclusive). This is encoded as \texttt{B-}, \texttt{I-}, \texttt{E-} tags of type \texttt{ORG}. Our model has predicted correct tags for the tokens ``First'' and ``Bank'', but annotates ``National'' as a miscellaneous span. This \texttt{I-MISC} tag is illegal. This tag should only follow \texttt{B-MISC} or \texttt{I-MISC}. The handling of these illegal tags can result in very different spans. In the original evaluation script from the CONLL 2000 shared task on noun-phrase chunking, a tag of a different type will trigger the ending of the previous entity and the start of a new one. This yields three entities, none of which match the gold entities. \citet{yang-etal-2018-design}, on the other hand, use a different entity resolution strategy where only the beginning and end tags are used. Illegal tags within the entity are ignored and under this scheme we get the correct entity. Mismatches in the entity decoding method, and specifically the handling of illegal transitions, can result in different entities based on the same tags. This renders results incomparable.
    }
    \label{tab:entity-resolution}
\end{table*}

\subsection{Entity Resolution}

Yet another problem area is the handling of malformed output sequences. As we established earlier, there are rules on the allowed transitions from one tag to another that are dictated by the span encoding scheme. With the data-driven modeling approaches that dominate these tasks, there is no guarantee that the output sequence will be well-formed. How these errors are handled can cause large differences in the output entities.

The evaluation script from the CONLL 2000 shared task \cite{tjong-kim-sang-buchholz-2000-introduction} on noun-phrase chunking, \texttt{conlleval.pl}, uses a policy that can be best described as: A difference in the types of tags triggers a shift in spans. This means that when you have a \texttt{B-PER} followed by an \texttt{I-LOC} you would create two spans, one for the person and one for the location, even though the location span did not legally start. The behavior is a clear outgrowth of the fact that this script originally was designed for IOB, but it is the de facto entity resolution policy.

In the NCRF++ toolkit, \citet{yang-etal-2018-design} use a different resolution strategy. As discussed in a Github issue\footnote{\href{https://github.com/jiesutd/NCRFpp/issues/87}{https://github.com/jiesutd/NCRFpp/issues/87}}, they only process legal spans. This means that they only look for \texttt{B-} tags to start entities and the corresponding \texttt{E-} tag to end it. They ignore changes in the type of intervening \texttt{I-} tags. As demonstrated in Table \ref{tab:entity-resolution}, this can produce very different entities when compared to the \texttt{conlleval.pl} outputs.

\section{Our Library}

To help avoid these kind of preventable mistakes, and to create a standardized policy on the creation of entities from illegal tag sequences, we introduce our library \textit{iobes}.

\subsection{Parsing}

\begin{listing}[t]
    \centering
    \begin{minted}{python}
    class Span(NamedTuple):
        type: str
        start: int
        end: int
        tokens: List[int]
    \end{minted}
    \caption{
        Our Span class. The value of the end attribute is one more than the index of the final token in the span. This formulation allows for recovery of the tokens in the span via Python list slicing.
    }
    \label{lst:span}
\end{listing}

Our library includes robust parsers for turning lists of token level annotations into spans, represented by the named tuple outlined in Listing \ref{lst:span}. Our library handles the IOB, BIO, IOBES, BILOU, and BMEWO schemes.

When encountering malformed token sequences, our library follows the \texttt{conlleval.pl} method of entity resolution where new entities are created when there is a difference in type between adjacent token labels. In addition to producing a list of spans, our library also creates a list of errors. These errors can help localize where illegal transitions are occurring. Common errors are things like switching entity types within a span, ending spans without an \texttt{E-} token, and staring spans with an \texttt{I-}.

\subsection{Conversion}

Our library also includes tooling for conversion between all of these different formats. When converting a malformed sequence of tokens, there is inherent uncertainty in what the true sequence of spans was. We discussed multiple entity resolution above, but we found that choosing one when converting malformed gold labels was overly prescriptive. Unlike the models predictions, we can have humans fix these malformed gold sequences. Rather than making some policy decision on the handling malformed sequences, we raise an exception and return the list of errors to help the user fix their gold labels. 

\subsection{Legal Transitions}

The span encoding formats dictate which tokens can follow others, for example, the IOBES scheme says the all entities must end with an \texttt{E-}, therefore an \texttt{O} cannot follow an \texttt{I-}. While most statistical models, especially those that have a global loss function, like the conditional random field (CRF) \cite{10.5555/645530.655813}, learn these relations, it is not guaranteed that these rules are followed.

Our library is able to enumerate the legality of all possible transitions. While models encode these rules as soft, learned constraints embedded in the model parameters, they can also be enforced by various techniques, like filtering sequences with illegal transitions or masking the scores these transitions get. By providing the legality of transitions, our library makes it easy to ensure a well-formed output.

\subsection{Engineering}

\begin{listing}[th]
    \centering
    \begin{minted}{python}
    class SpanFormat:
        BEGIN: str
        INSIDE: str
        END: str
        SINGLE: str
    \end{minted}
    \caption{
        Our span encoding data structure. These classes allow us to reuse the same parsing code with different classes to parse IOBES, BILOU, and BMEWO labels. By checking token prefixes against values in this data structure, instead of explicit checks against strings like \texttt{"S-"} or \texttt{"E-"}, we can use the same code for all formats. Having only a single function means there is less surface area for bugs to creep in and allows us to test it much more thoroughly.
    }
    \label{lst:token-scheme}
\end{listing}

Our main goal in this library is correctness. We achieve that by reusing code as much as possible. By defining data structures that contain the special prefixes each encoding scheme uses, like the one in Listing \ref{lst:token-scheme}, we can use a single, well-tested function to perform some action---such as span parsing or converting a span into token labels---for multiple encoding schemes. We also use property based tests and the fuzzing of inputs to ensure our code is behaving properly. Our tests are automatically run via CI/CD on multiple operating systems to ensure a smooth experience across platforms.

Our library is lightweight and has no dependencies. Keeping it small makes it as painless as possible to integrate with unique workflows.

In order to handle multiple encoding schemes, we provide both specific functions like \texttt{parse\_spans\_iob}, as well as functions like \texttt{parse\_spans}, that dispatch on the value of the \texttt{span\_type} parameter. Similarly, the legality of various transitions is available in multiple formats, including a mask that is ready to be applied to a CRF. We provide multiple interfaces like this to support as many use cases as possible.

\section{Conclusion}

Many span level tasks in natural language processing are recast as token-level labeling. There are many encoding schemes used to convert these tokens into spans and these schemes dictate which tokens can follow other ones. Unfortunately, processing these tokens is a common place where errors and mistakes manifest. We have shown how mistakes in conversion code, gold annotations, and entity resolution, in the presence of malformed tag sequences, render results incomparable. 

To remedy these problems, we introduce \textit{iobes}, a small, well-tested Python library that helps standardize the processing of spans. Our library helps with parsing token labels into a list of spans, identifying locations of errors in token sequences, converting between span encoding schemes, and enumerating which transitions are allowed and which are not. Our library will help avoid these errors and will ensure that results created by different researchers are comparable.

\bibliography{emnlp2020}
\bibliographystyle{acl_natbib}

%\appendix

\end{document}